\documentclass{amsart}

\usepackage{algorithm}
\usepackage[noend]{algorithmic}

\usepackage{amssymb}
\usepackage{multirow}

\usepackage[final]{graphicx}
\usepackage{latexsym}

\usepackage{amsmath}
\newcommand{\argmax}{\operatornamewithlimits{arg\,max}} 

\begin{document}
\title[Ensemble Relational Learning based on\\ Selective Propositionalization]{Ensemble Relational Learning based on\\ Selective Propositionalization}

\author[N. Di Mauro]{Nicola Di Mauro}
\address[Nicola Di Mauro]{Department of Computer Science, LACAM Laboratory, University of Bari
``Aldo Moro'', Bari, Italy}
\email{nicola.dimauro@uniba.it}

\author[F. Esposito]{Floriana Esposito}
\address[Floriana Esposito]{Department of Computer Science, LACAM Laboratory, University of Bari ``Aldo Moro'', Bari,
Italy}
\email{floriana.esposito@uniba.it}

\maketitle

\begin{abstract}  Dealing with structured data needs the use of expressive representation formalisms
that, however, puts the problem to deal with the computational complexity of the machine learning
process. Furthermore, real world domains require tools able to manage their typical uncertainty.
Many statistical relational learning approaches try to deal with these problems by combining the
construction of relevant relational features with a probabilistic tool.  When the combination is
static (\emph{static propositionalization}), the constructed features are considered as boolean
features and used offline as input to a statistical learner; while, when the combination is dynamic
(\emph{dynamic propositionalization}), the feature construction and probabilistic tool are combined
into a single process.   
In this paper we propose a \emph{selective propositionalization}
method that search the optimal set of relational features to be used by a probabilistic learner in
order to minimize a loss function.  The new propositionalization approach has been combined with the
random subspace ensemble method.  Experiments on real-world datasets shows the validity of the
proposed method. 
\end{abstract}

\section{Introduction}
\label{sec:Introduction}
Dealing  with  relational domains  requires  the use  of  expressive  and structured  representation
formalisms  such as  graphs  or  first-order logic  already  used in  the  area  of Inductive  Logic
Programming  (ILP)~\cite{DBLP:journals/jlp/MuggletonR94}. Furthermore,  real  world domains  require
tools  able  to manage  their  typical  uncertainty. Frameworks  and  systems  able  to both  manage
relational descriptions  and to reason in probabilistic  way have been
emerged in the research area of
Probabilistic   Inductive  Logic   Programming  (PILP)~\cite{DBLP:conf/ilp/2008p}   and  Statistical
Relational Learning (SRL)~\cite{1296231}. 

Most of the ILP learning approaches build models by searching (constructing) for
\emph{good}    relational    features   guided    by    a    scoring    function,   such    as    in
\texttt{FOIL}~\cite{DBLP:journals/ml/Quinlan90}. 
In the classical ILP  setting the constructed features are assumed to  be as strong constraints that
the observations  must fulfill. Typically, ILP  algorithms search Horn
clauses that must cover or subsume all the positive observations an no
negative ones. In order to soft this assumption, in many PILP and SRL systems 
this feature construction process is combined with a discriminative/generative 
probabilistic method in order to deal with noisy data and uncertainty, such as in
\texttt{kFOIL}~\cite{DBLP:conf/aaai/LandwehrPRF06}
 and Markov Logic Networks
(MLNs)~\cite{Richardson06}. The combination may be
\emph{static} or \emph{dynamic}. In the former case, named \emph{static propositionalization}, the
constructed features are usually considered 
as boolean features and used offline as input to a propositional statistical learner; while in the
latter  case,   named  \emph{dynamic  propositionalization},   the  feature  construction   and  the
probabilistic model selection are combined into a single process~\cite{Kramer:2001:PAR:567222.567236}.  

In this  paper we propose a \emph{selective  propositionalization} approach for the  general case of
relational  learning, originally  presented  in \cite{ndm11ismis}  for
relational sequences only, that search the
optimal set of relational features to be used by a probabilistic learner in order to minimize a loss
function. In  particular, after a  first feature  construction phase, the  set of the  most relevant
features minimizing  a Bayesian  classifier's probability error  are stochastically  searched. Here,
after  an improved  formalization  of  the proposed  approach  and its  evaluation  in  the area  of
relational  propositionalization, the first  aim is  to investigate  whether the  resulting proposed
method is a valuable tool when applied to classical relational domains. 
Furthermore, in this paper the new propositionalization approach has been combined with the \emph{random
  subspace ensemble  method} (RSM)~\cite{Ho:1998:RSM:284980.284986} trying  to improve the  generalization accuracy of a  single base
classifier.  Experiments on real-world datasets, when compared to some state of the art SRL systems,
show the validity of the proposed methods.  

After providing some motivations and related works in the next section, Section~\ref{sec:Lynx} describes the
proposed selective  propositionalization approach and  its combination
with the RSM, while 
Section~\ref{sec:exp} provides a qualitative validation of the approach. Finally, Section~\ref{sec:conc}
concludes the paper.

\section{Motivation and Related Work}
Traditional relational learning approaches dynamically generate features providing information about
observations, interleaving the feature construction and the model construction. A way to tackle the task
of  inferring predictive  and discriminant  features in  relational learning  is to  reformulate the
problem     into    an     attribute-value     form    and     then     apply    a     propositional
learner~\cite{Kramer:2001:PAR:567222.567236}.  

\texttt{nFOIL}~\cite{DBLP:conf/aaai/LandwehrKR05}   and  \texttt{kFOIL}~\cite{DBLP:conf/aaai/LandwehrPRF06}   are  two
examples  of dynamic propositionalization.  Differently from  the static  propositionalization, where
firstly the  features have  been generated  and then the  parameters for  a statistical  learner are
estimated, they tightly  integrates the learning of the features  with the statistical propositional
learner.  The criterion  according to  which the  features are  generated is  that of  a statistical
learner, a na\"{i}ve Bayes in the case of \texttt{nFOIL} and a support vector machine (SVM) for \texttt{kFOIL}.
Both      the     methods     employ      an     adaptation      of     the      well-known     \texttt{FOIL}
algorithm~\cite{DBLP:journals/ml/Quinlan90}  that implements  a  separate-and-conquer rule  learning
algorithm. 

The   generic  \texttt{FOIL}   algorithm   iteratively  searches   for
relational features (i.e., clauses) that score well with respect
to the observations and the current hypothesis and adds them to the current hypothesis. Each feature
is greedily searched by using a general-to-specific hill-climbing search strategy. 
The adaption of this algorithm to the case  of \texttt{nFOIL} and \texttt{kFOIL} corresponds to evaluate the candidate
features according to a probabilistic scoring function.

This approach is however sensitive to the ordering of the selected candidate features that determine
the choice of the following features. 
Furthermore, for  the case of na\"{i}ve  Bayes, as reported  in \cite{DBLP:conf/uai/LangleyS94}, the
model can suffer from oversensitivity to redundant and/or irrelevant attributes. Even for the SVMs 
has been shown  in~\cite{DBLP:conf/nips/WestonMCPPV00} that they can perform  badly in the situation
of many irrelevant examples and/or features. 

Since, the effectiveness of learning algorithms
strongly  depends on the  used features,  a feature  selection task  is very  desirable. The  aim of
feature selection is to find an optimal  subset of the input features leading to high classification
performance, or,  more generally, to carry  out the  classification task optimally.  However, the
search  for a  variable subset  is a  NP-hard  problem. Therefore,  the optimal  solution cannot  be
guaranteed to be  reached except when performing  an exhaustive search in the  solution space. Using
stochastic local search  procedures~\cite{sls04} allows one to obtain  good solutions without having
to explore the whole solution space.

Differently from a dynamic propositionalization, we firstly  construct a set of features and then we
adopt a wrapper feature selection approach, that uses a stochastic local search procedure, embedding
a na\"ive Bayes classifier to select an  optimal subset of  the features.  The
optimal subset is  searched using a Greedy Randomized Search  Procedure (GRASP)~\cite{feo95} and the
search is  guided by  the predictive power  of the  selected subset computed  using a  na\"ive Bayes
approach.  

In particular, given a training dataset $\mathcal D= \{\mathbf{x}_i,c_i\}_{i=1}^n$ of $n$ relational examples,
characterized by a  set of $m$ relational features  $ X = \{f_i\}_{i=1}^m$, and a target discrete
random variable $c$, generating class labels $c_i$, the aim of this
paper is to find a subset of $ X$ that optimally characterizes the variable $c$ minimizing
the classifier's probability error. 

\section{\texttt{Lynx} and \texttt{Lynx-RSM}}
\label{sec:Lynx}
This section  reports the components  of the \texttt{Lynx}  system and
its ensemble extension \texttt{Lynx-RSM},  implementing  a  probabilistic
relational classifier. Specifically, we start to report their feature
construction  capability  and  the  adopted  relational  feature-based
classification model,  as already defined  for the case  of relational
sequence  learning~\cite{ndm11ismis}.  Here   the  approach  has  been
generalized to  the case of  relational learning and then  extended to
the case of ensemble learning.

\subsection{Relational Feature Construction}
\label{sec:construction}
The first  step of \texttt{Lynx}  carries out a  feature construction process by mining
frequent Prolog queries (relational features) adopting an approach similar to that reported in~\cite{Kramer01}.
 The   algorithm for frequent  relational query mining is based on  the same
idea as the  generic  level-wise search  method, performing a  breadth-first search  in the  lattice
of queries  ordered  by a specialization relation $\preceq$. 
The algorithm starts with
the most general Prolog queries. 
At  each step it tries to  specialize all the candidate frequent queries, discarding
the non-frequent ones and storing those whose length is  less or equal to a  user specified input
parameter.  Furthermore,  for  each  new refined  query,  semantically  equivalent
patterns are detected, by using the $\theta_{\mathrm{OI}}$-subsumption relation~\cite{ferilli03ismis}, 
and discarded. In  the
specialization  phase the  specialization  operator, basically,  adds atoms to the query. 

Now, having a set of relational features, we need a way to use them in
order to  correctly classify unseen examples. 
Given the training set $\mathcal D= \{X_i,c_i\}_{i=1}^n$ of $n$ relational examples, where
$c$ denotes the discrete class random variables taking values from  $\{1,2,\ldots,Q\}$,
%
the goal is to learn a function $h  : x \rightarrow c$ from $\mathcal D$ that  predicts the label for each
unseen observation.  

 Let $\mathcal  Q$, with  $|\mathcal Q| =d$,  be the  set of  features
obtained in  the first  step of the  \texttt{Lynx} system  (the queries mined  from $\mathcal D$).
For each
example $X_k$ we  can build a $d$-component  vector-valued $\mathbf x_k  = (x_k^1, x_k^2,
\ldots, x_k^d)$ random variable  where each $x_k^i \in \mathbf x_k$ is 1 if  the query $q_i \in \mathcal
Q$ subsumes example $\mathbf x_k$, and 0 otherwise, for each $1 \leq i
\leq   d$.   This  exactly   corresponds  to   a  propositionalization
process.   The   relational   observations   are  transformed   to   a
propositional form on which classical statistical learner may be applied.

Using the Bayes' theorem,  if $p(c_j)$ describes the prior probability of class $c_j$, then the
posterior   probability  $p(c_j|   \mathbf   x)$  can   be   computed  from   $p(\mathbf  x|   c_j)$
as
\begin{equation}
p(c_j|\mathbf  x)  =   \frac{p(\mathbf  x  |  c_j)p(c_j)}{\sum_{i=1}^Q
  p(\mathbf x|c_i)p(c_i)}.
\end{equation} 

Given a set of discriminant functions $\{g_i(\mathbf x)\}_{i=1}^Q$, a classifier is said to
assign the  vector $\mathbf x$ to class $c_j$ if $g_j(\mathbf  x) > g_i(\mathbf x)$  for all $j
\neq  i$.   Taking  $g_i(\mathbf x)  =  P(c_i  |  \mathbf  x)$,  the maximum  discriminant  function
corresponds  to  the  \emph{maximum  a  posteriori}  (MAP)  probability.   For  minimum  error  rate
classification, the following discriminant function will be used: 
\begin{equation}
g_i(\mathbf x) = \ln p(\mathbf x | c_i) + \ln p(c_i).
\end{equation}
Given $\mathbf x = (x_1, \ldots, x_d)$, we
define    
$p_{ij}=\mathrm{Prob}(x_i=1|c_j)$    
with    the
components of $\mathbf x$ being statistically independent for all $x_i \in \mathbf x$. 
%
The estimator $\hat{p}_{ij}$ of the factor $p_{ij}$ corresponds to the frequency counts on the
training observations: $$\hat{p}_{ij} = \eta_{i,j}(\mathcal D, \mathcal Q) = |\{ X_k, c_k \in \mathcal D
| c_k = j \wedge q_i \in \mathcal Q\ \mathrm{subsumes}\  X_k\}| / \eta_j(\mathcal D),$$ where
$\eta_j(\mathcal D) = |\{ X_k, c_k \in \mathcal D | c_k = j|$. The estimator $\hat p(c_j)$
of $p(c_j)$ is $\eta_j(\mathcal D) / |\mathcal D|$. 
By assuming  conditional independence 
$p(\mathbf x| c_j) = \prod_{i=1}^d (p_{ij})^{x_i}(1 - p_{ij})^{1-x_i}$, yielding the discriminant function
\begin{equation}
g_j(\mathbf x) =  \ln p(\mathbf x | c_j) + \ln p(c_j)=
 \sum_{i=1}^d x_i \ln \frac{ p_{ij}}{1 - p_{ij}} + \sum_{i=1}^d \ln (1 - p_{ij})  + \ln p(c_j).
\label{df}
\end{equation}
The minimum probability  error is achieved by deciding $c_k$
if $g_k(\mathbf x) \geq g_j(\mathbf x)$ for all $j$ and $k$.

\subsubsection{Feature Selection with Stochastic Local Search}
After having constructed a set of features, and presented a method to use those features to classify
unseen sequences, now the  problem is how to find a subset  of these features that optimizes
the prediction accuracy. 
The optimization problem of selecting a subset of features  with
a  superior classification  performance  may  be formulated  as  follows. Let  $\mathcal  P$ be  the
constructed  original set  of features,  and let  $f  : 2^{|\mathcal  P|} \rightarrow  \mathbb R$ be a
function scoring a selected subset $X \subseteq  \mathcal P$. The problem of feature selection is to
find  a subset  $\widehat{X} \subseteq  \mathcal P$  such that  $f(\widehat{X}) =  \max_{Z \subseteq
  \mathcal P} f(Z)$.
An  exhaustive approach  to this  problem would  require examining  all $2^{|\mathcal  P|}$ possible
subsets of  the feature set $\mathcal  P$, making it impractical  for even small values of $|\mathcal
P|$.  The  use of  a stochastic  local search procedure~\cite{sls04} allows to obtain  \emph{good} solutions
without having to explore the whole solution space. 


Given a subset $P \subseteq \mathcal P$, for each observation $ X_j \in \mathcal D$ we let the
classifier find the MAP hypothesis $\widehat{h}_P(  X_j) = \argmax_{i} g_i(\mathbf x_j)$ by adopting the
discriminant  function reported in  Eq.~\ref{df}, where  $\mathbf x_j$  is the  feature based
representation of the observation $X_j$ obtained using the features in $P$. The initial optimization
problem corresponds to minimize the expectation $\mathbb E[\mathbf 1_{\widehat{h}_P(X_j) \neq c_j}]$
where $\mathbf 1_{\widehat{h}_P(X_j)  \neq c_j}$ is the characteristic
function of the training
observation $X_j$ returning $1$  if $\widehat{h}_P(X_j) \neq c_j$, and
$0$ otherwise (i.e., the 0-1 loss function).
Finally, given $\mathcal D$ the training set with $|\mathcal D|=m$ and $P$ a set of features, 
the number of classification errors made by the Bayesian model is
\begin{equation}
  err_D(P)  = m  \mathbb E  \left [\mathbf  1_{\widehat{h}_P(X_j) \neq
      c_j} \right ].
\label{eq:eval}
\end{equation}

\label{sec:grasp}

Consider  a \emph{combinatorial optimization}  problem, where  one is  given a  discrete set  $X$ of
solutions and an objective function $f : X \rightarrow \mathbb R$ to be minimized, and seek
a solution $x^* \in X$ such that $\forall x \in X: f(x^*) \leq f(x)$. 
A  method to  find  high-quality solutions  for  a combinatorial  problem consists of a  two-step  approach
made up of   a    greedy   construction    phase    followed   by    a   perturbative    local
search~\cite{sls04}.  

The greedy  construction  method starts  the  process  from an  empty
candidate solution  and at  each construction  step adds the  best ranked  component according  to a
heuristic  selection  function. Then,  a  perturbative local  search  algorithm,  searching a  local
\emph{neighborhood}, is  used to improve the  candidate solution thus obtained.   Advantages of this
search method are a much better solution quality and fewer perturbative improvement steps needed to reach
the local optimum.  

\texttt{GRASP}~\cite{feo95} solves the  problem of the limited
number  of different  candidate  solutions generated  by  a greedy  construction  search method  by
randomizing the  construction method.  \texttt{GRASP} is an  iterative process combining at  each iteration a
construction and a local  search phase. In the construction phase a  feasible solution is built, and
then its neighbourhood is explored by the local search. 

Algorithm~\ref{alg:grasp} reports the \texttt{GRASP$^{\mathtt{FS}}$} procedure included in the \texttt{Lynx}
system to  perform  the feature
selection task.  In each iteration, it computes a  solution $S \in \mathcal S$ by using a randomized
constructive search procedure and then applies a  local search procedure to $S$ yielding an improved
solution. The main procedure  is made up of two components: a constructive  phase and a local search
phase.  

\begin{algorithm}
\caption{\texttt{GRASP$^{\mathtt{FS}}$}}
\label{alg:grasp}
\begin{algorithmic}[1]
\REQUIRE{$\mathcal D$: the training set;\\ $\mathcal F$: a set of $n$
  relational features;\\ \emph{maxiter}: maximum
  number of iterations;\\ $err_{\mathcal D}(\mathcal F)$: the evaluation function (see Eq.~\ref{eq:eval})}
\ENSURE{an ordered set of $m$ solutions $\mathcal S = \{ S_i | S_i \subseteq \mathcal F\}_{i=1,\ldots,m}$}
\STATE $\mathcal S = \emptyset$ 
\STATE iter $= 0$
\WHILE{iter $<$ maxiter}
  \STATE $\alpha = $ rand(0,1)
  \STATE $S = \emptyset$; $i = 0$; improved = true
\STATE $err_{\mathcal D}(S) = +\infty$
  \WHILE{$i < n$ and improved} 
  \STATE $\mathcal C = \{ C | C = add(S,A)\}$ for each component $A \in \mathcal F$ s.t. $A
  \not\in S$
    \STATE $\overline{s} = \max_T \{err_{\mathcal D}(C) | C \in \mathcal C\}$
    \STATE $\underline{s} = \min_T \{err_{\mathcal D}(C) | C \in \mathcal C\}$
    \STATE RCL $= \{C \in \mathcal C | err_{\mathcal D}(C) \leq \underline{s} + \alpha (\overline{s} - \underline{s})\}$
    \STATE select a candidate solution $C$, at random, from the set RCL
    \IF{$err_{\mathcal D}(C) < err_{\mathcal D}(S)$}
    \STATE $S = C$
    \ELSE
    \STATE improved = \textbf{false}
    \ENDIF
    \STATE $i \leftarrow i + 1$
  \ENDWHILE
  \STATE  $\mathcal N  = \{  N \in  neigh(S) |  err_{\mathcal  D}(N) <
  err_{\mathcal D}(S)\}$
  \WHILE{$\mathcal N \neq \emptyset$}
    \STATE select a new solution $S \in \mathcal N$
  \STATE  $\mathcal N  = \{  N \in  neigh(S) |  err_{\mathcal  D}(N) <  err_{\mathcal D}(S)\}$
  \ENDWHILE
  \IF{$err_{\mathcal  D}(S) <  \min  \{err_{\mathcal D}(S')  | S'  \in
    \mathcal S\}$}
  \STATE add $S$ to $\mathcal S$
  \ENDIF
  \STATE iter = iter + 1
\ENDWHILE
\STATE \textbf{return} $\mathcal S$
\end{algorithmic}
\end{algorithm}

The constructive search algorithm (lines 4-12) used in \texttt{GRASP$^{\mathtt{FS}}$} iteratively adds a solution
component  by  randomly selecting  it,  according  to a  uniform  distribution,  from  a set,  named
\emph{restricted  candidate list}  (RCL), of  highly ranked  solution components  with respect  to a
greedy function $g : \mathcal S \rightarrow \mathbb R$.  

The  probabilistic  component  of
\texttt{GRASP$^{\mathtt{FS}}$}  is characterized  by a random choice of one of  the best  candidates in  the
RCL. In  our case the greedy  function $g$ corresponds  to the error function  $err_D(P)$ previously
reported  in  Eq.~\ref{eq:eval}.  In  particular,  given $err_D(P)$,  the  heuristic  function,  and
$\mathcal S$, the set of feasible solutions, $\underline{s} = \min \{ err_D(S) | S \in \mathcal S\}$
and $\overline{s} = \max \{ err_D(S) | S \in  \mathcal S\}$ are computed. Then the RCL is defined by
including in it all the components $S$ such that $err_D(S) \geq \underline{s} + \alpha (\overline{s}
- \underline{s})$.  

To improve the solution generated by the construction phase, a local search is used (lines 13-16).  It works by
iteratively replacing the current solution with a better solution taken from the neighborhood of the
current solution while such a better solution exists.
Given $\mathcal P$ the  set of patterns, in order to build  the neighborhood \emph{neigh(S)} of a
solution $S = \{p_1, p_2, \ldots, p_t\} \subseteq \mathcal P$,   the following
operators are exploited:
\begin{description}
\item[\textbf{add}:] $S \rightarrow S \cup \{p_i\}$ where $p_i \in \mathcal P \setminus S$;
\item[\textbf{replace}:] $S \rightarrow S \setminus \{p_i\} \cup \{p_k\}$ where
  $p_i \in S$ and $p_k \in \mathcal P \setminus S$.
\end{description}
In particular, given a  solution $S \in \mathcal S$, the elements  of the neighborhood $neigh(S)$ of
$S$ are those solutions that can be  obtained by applying an elementary modification (add or replace)
to $S$. Local search starts from an  initial solution $S^0 \in \mathcal S$ and iteratively generates
a series of improving solutions $S^1, S^2, \ldots$.  At the $k$-th 
iteration,  $neigh(S^k)$  is searched  for  an  improved solution  $S^{k+1}$  such  that $err_D(S^{k+1})  <
err_D(S^{k})$. If  such a solution is  found, it becomes the current solution. Otherwise,  the search ends
with $S^{k}$ as a local optimum. 
After each iteration,  the given solution is added  to the ordered set
of solutions $\mathcal S$.
The algorithm  does not  return the best  local solution, but  all the
found ones that can then be used by the following ensemble algorithm.

\subsection{Ensemble Learning}
Combining the  predictions of multiple classifiers,  known as ensemble
learning~\cite{Rokach:2010:EC:1713727.1713730}, is one of the 
standard  and  most  important technique  for  improving  the  classification accuracy  in  machine
learning. While bagging  and boosting works on sampling the  training observations, other techniques
investigate  the   performance of  classifier ensembles  trained using  attribute  subsets, where
selecting the optimal subsets of relevant features plays an important role.

One popular  ensemble method is  the random subspace  method (RSM)~\cite{Ho:1998:RSM:284980.284986},
whose idea is to use a sample of the feature set for each classifier in the ensemble.
Then the ensemble operates by taking the majority vote of a predefined number of classifiers.

Assuming that each observation is defined on a $p$-dimensional vector, described by $p$ features.
The  RSM  randomly  selects $r  <  p$  features  from  the  $p$-dimensional data  set,  obtaining  a
$r$-dimensional random  subspace of the  original $p$-dimensional feature  space. Given $m$  of such
random subspaces,  a classifier is learned  for each subspace and  then they are  combined by simple
majority voting or by averaging the conditional probability of each class. The parameters of the RSM
are the ensemble size $m$ and the cardinality $r$ of the feature subset.

Here  the approach  we  used to  construct  the ensemble  is slightly  different  form the  original
one.  Given the  set of  the constructed  relational  feature, each  iteration of  the feature  selection
algorithm (see Algorithm~\ref{alg:grasp}) gives us a random subspace to be considered to build the classification model.  In
particular, we avoid to set the parameter $r$, since it may change for
each iteration of the \texttt{GRASP$^{\mathtt{FS}}$} procedure 
governed by the  stopping condition.  Hence fixing  the size  of the  ensemble the  Algorithm~\ref{alg:lynx-rsm}
reports the procedure we used to build the ensemble.

\begin{algorithm}
\caption{\texttt{lynx-RSM}}
\label{alg:lynx-rsm}
\begin{algorithmic}[1]
\REQUIRE{$\mathcal D$: the training set;\\
\emph{maxiter}: maximum number of iterations of the \texttt{GRASP$^{\mathtt{FS}}$} procedure;\\
 $err_{\mathcal D}(\mathcal F)$: the evaluation function (see Eq.~\ref{eq:eval});\\
$m$: the number of individual classifiers in the ensemble}
\STATE build the set $\mathcal F$ of $n$ relational features
\STATE $\mathcal S$ = \texttt{GRASP$^{\mathtt{FS}}$}($\mathcal D, \mathcal F, maxiter,
m , err_{\mathcal D}$)
\STATE train the classifier for each random subspace $S_i \in
\mathcal S$
\STATE  for classifying a  new observation,  combine the  predictions of  the $m$
individual  classifiers  by   combining  the  posterior  probabilities
$p(c_j| \mathbf x_j)$ of each one
\end{algorithmic}
\end{algorithm}

We firstly  build the set $\mathcal  F$ of relational  features and we
call the  \texttt{GRASP$^{\mathtt{FS}}$} procedure in  order to obtain
an ordered set $\mathcal S$ of $m$ random subspaces. Then we train a
classifier  for  each  of  these  subspaces  and  we  classify  a  new
observation by combining the predictions of the $m$
individual  classifiers  by   combining  the  posterior  probabilities
$p(c_j| \mathbf x_j)$ of each classifier.

\section{Experiments}
\label{sec:exp}

We  tested  the proposed  \texttt{Lynx}  approach,  and its  extension
\texttt{Lynx-RSM}, on two well known ILP datasets, the Mutagenesis and
the   Alzheimer  datasets,  and   on  the   widely  used   UW-CSE  SRL
dataset~\cite{Singla:2005:DTM:1619410.1619472}.

The Mutagenesis dataset~\cite{Srinivasan:1996:TMS:241085.241105} regards the problem to predict the mutagenicity of a
set of molecules based on their chemical structure. Of the 188 molecules, 125
have positive log mutagenicity whereas 63 molecules have zero or negative log
mutagenicity. The molecules with positive log mutagenicity are labeled active
and the remaining are labeled inactive.
As in~\cite{DBLP:conf/aaai/LandwehrPRF06} we used atom and bond information
only. 

In the Alzheimer dataset~\cite{king95} the aim is to compare
analogues of Tacrine, a drug against Alzheimer’s disease, according to four desirable properties: inhibit amine re-uptake, low toxicity, high acetyl cholinesterase inhibition, and good
reversal of scopolamine-induced memory deficiency. Examples consist of pairs  of
two analogues,  and are labeled positive  if and only if  the first is
rated higher than the second with regard
to the property of interest. 

The UW-CSE dataset~\cite{Singla:2005:DTM:1619410.1619472}, widely used
in SRL,
regards the Department of Computer Science and Engineering at
the University of Washington, describing relationships among professors, students, courses and
publications with 3212 true ground atoms over 12 predicates.
The task is to predict the relationship \texttt{advisedBy(X,Y)} using in turn four of the five
research areas (ai, graphics, language, theory and systems) for training and the remaining one for
testing as in~\cite{Singla:2005:DTM:1619410.1619472}.

For the  Mutagenesis and Alzheimer datasets, we  compared our proposed
approach to \texttt{nFOIL} and \texttt{kFOIL},  while for the UW-CSE dataset we compared
\texttt{Lynx}    to     two    systems    learning     Markov    Logic
Networks~\cite{Richardson06} (MLNs), specifically 
\texttt{LSM}~\cite{DBLP:conf/icml/KokD10} (Learning Using Structural Motifs) and
\texttt{LHL}~\cite{Kok:2009:LML:1553374.1553440}     (Learning     via
Hypergraph Lifting). 
MLNs are one of the most important representation formalism in SRL combining the
expressiveness of first-order logic with the robustness of probabilistic representations. An MLN is a
set of weighted first-order formulas, and learning its
structure consists of learning both formulas and their
weights. \texttt{LSM} and \texttt{LHL} are two state of the art MLNs structure learning algorithms.

\begin{table}
\centering
\begin{tabular}{|c|c|c|c|c|}
\hline
 & \texttt{Lynx} & \texttt{Lynx-RSM} & \texttt{kFOIL} & \texttt{nFOIL} \\
\hline \hline
Mutagenesis & 86.7 $\pm$ 6.2 & \textbf{89.4} $\pm$ 7.4 & 77.7 $\pm$ 14.5 & 75.4 $\pm$ 12.3 \\
\hline
\hline
Alzheimer amine & 88.6 $\pm$ 5.8 & 88.8 $\pm$ 5.5 & 89.8 $\pm$ 5.7 & 86.3 $\pm$ 4.3 \\
Alzheimer toxic & 94.1 $\pm$ 2.9 & 94.4 $\pm$ 2.1 & 90.0 $\pm$ 3.8 & 89.2 $\pm$ 3.4 \\
Alzheimer acetyl & 89.4 $\pm$ 3.3 & 89.9 $\pm$ 1.7 & 90.6 $\pm$ 3.4 & 81.2 $\pm$ 5.2 \\
Alzheimer memory & 82.1 $\pm$ 6.8 & 82.9 $\pm$ 4.6 & 80.5 $\pm$ 6.2 & 72.9 $\pm$ 4.3 \\
\hline
Alzheimer mean & 88.6 $\pm$ 4.7 & \textbf{89.0} $\pm$ 3.5 & 87.7 $\pm$ 4.8 & 82.4 $\pm$ 4.3 \\
\hline
\end{tabular}
\caption{Experimental  results   on  the  Mutagenesis   and  Alzheimer
  datasets.}
\label{res:mut}
\end{table}

In all the experiments, 
the maximum length parameter for the relational features learned by
\texttt{Lynx} has  been set  to 6, while  the feature  selection grasp
procedure has been iterated 100 times.  
Table~\ref{res:mut}  reports the experimental  results with  a 10-fold
cross validation on both the
Mutagenesis  and Alzheimer datasets  obtained with  \texttt{Lynx} when
compared to \texttt{nFOIL} and \texttt{kFOIL}~\cite{DBLP:journals/ml/LandwehrPRF10}. For both  \texttt{nFOIL} and
\texttt{kFOIL}   we  used   the   same  parameters   as  reported   in
\cite{DBLP:journals/ml/LandwehrPRF10}.  The first  column  reports the
accuracy obtained  by \texttt{Lynx} that  is greater when  compared to
that obtained by \texttt{nFOIL}  and \texttt{kFOIL}. The second column
reports the  accuracy of \texttt{Lynx-RSM},  obtained with an
ensemble  of  40  classifiers  for  the  Mutagenesis  dataset  and  12
classifiers for the Alzheimer dataset, that as we can see is always greater
than  that  obtained  with   \texttt{Lynx}  that  use  a  single  base
classifier. This  first results confirms the improvements  that can be
obtained with an ensemble approach. 

In  the  second experiment  we  compared  the  proposed approach  with
respect  to a classical  SRL formalism.  Table~\ref{tab:uwcse} reports
the  Area   under  the  ROC  and  Precision-Recall   (PR)  curves  for
\texttt{Lynx} and its  ensemble extension. As we can  see, the results
of \texttt{Lynx-RSM},  obtained  using an  ensemble of  90
classifiers, are  better than that obtained with  a single classifier,
already confirming the validity of the proposed approach.

\begin{table}
  \centering
  \begin{tabular}{l|c|c||c|c|}
    \cline{2-5}
    \multirow{2}{*}{} &  \multicolumn{2}{c||}{\texttt{Lynx}} & \multicolumn{2}{c|}{\texttt{Lynx-RSM}}\\ 
    \cline{2-5}  & AUC ROC & AUC PR & AUC ROC & AUC PR\\
    \hline \hline
    \multicolumn{1}{|l|}{ai} & 0.948 & 0.080 & 0.957 & 0.125\\
    \multicolumn{1}{|l|}{graphics} & 0.977 & 0.106 & 0.990 & 0.388\\
    \multicolumn{1}{|l|}{language} & 0.985 & 0.396 & 0.990 & 0.447\\
    \multicolumn{1}{|l|}{systems} & 0.949 & 0.059 & 0.965 & 0.105\\
    \multicolumn{1}{|l|}{theory} & 0.961 & 0.162 & 0.973 & 0.184\\ \hline
    \multicolumn{1}{|l|}{\textbf{mean}} & 0.964 & 0.161 & \textbf{0.975} & \textbf{0.250} \\ \hline
  \end{tabular}
  \caption{Area under the curve for ROC and PR on the UW-CSE
  dataset for  \texttt{Lynx} and \texttt{Lynx-RSM}.}
  \label{tab:uwcse}
\end{table}

Finally,  Table  \ref{exp:uw}  reports  the results  of  our  proposed
approach  when  compared  to  that  obtained  with  the  \texttt{LHL}  and  \texttt{LSM}
systems. For  \texttt{LHL} and \texttt{LSM} we  used the same parameters  as reported in
\cite{DBLP:conf/icml/KokD10}.  As  we can see,  both \texttt{Lynx} and
\texttt{Lynx-RSM}  outperform  \texttt{LHL}  and  \texttt{LSM},  thus  confirming  their
validity.

\begin{table}
\centering
\begin{tabular}{|c|c|c|}
\hline
&  ROC & PR\\
\hline \hline
Lynx & 0.964 $\pm$ 0.016 & 0.161 $\pm$ 0.137\\
Lynx-RSM& \textbf{0.975} $\pm$ 0.015 & \textbf{0.250} $\pm$ 0.157\\
\hline
\texttt{LHL} &  0.549 $\pm$ 0.079 & 0.010 $\pm$ 0.005\\
\texttt{LSM} &  0.870 $\pm$ 0.036 & 0.040 $\pm$ 0.023\\
\hline
\end{tabular}
\caption{Area under the curve for ROC and PR on the UW-CSE
  dataset for \texttt{Lynx}, \texttt{Lynx-RSM}, \texttt{LHL} and \texttt{LSM}.}
\label{exp:uw}
\end{table}

\section{Conclusion}
\label{sec:conc}
Dealing with structured data needs the use of expressive representation formalisms
that, however, puts the problem to deal with the computational complexity of the machine learning
process. Furthermore, real world domains require tools able to manage their typical uncertainty.
In this paper we proposed a selective propositionalization
method that search the optimal set of relational features to be used by a probabilistic learner in
order to minimize a loss function.  The new propositionalization approach has been combined with the
random subspace ensemble method.  Experimental results on real-world datasets shows the validity of the
proposed method when compared to other SLR approaches.

\bibliographystyle{splncs}
\bibliography{biblio}

\begin{thebibliography}{10}

\bibitem{DBLP:journals/jlp/MuggletonR94}
Muggleton, S., Raedt, L.D.:
\newblock Inductive logic programming: Theory and methods.
\newblock J. Log. Program. \textbf{19/20} (1994)  629--679

\bibitem{DBLP:conf/ilp/2008p}
De~Raedt, L., Frasconi, P., Kersting, K., Muggleton, S., eds.:
\newblock Probabilistic Inductive Logic Programming - Theory and Applications.
\newblock Volume 4911 of LNCS., Springer (2008)

\bibitem{1296231}
Getoor, L., Taskar, B.:
\newblock Introduction to Statistical Relational Learning (Adaptive Computation
  and Machine Learning).
\newblock The MIT Press (2007)

\bibitem{DBLP:journals/ml/Quinlan90}
Quinlan, J.R.:
\newblock Learning logical definitions from relations.
\newblock Machine Learning \textbf{5} (1990)  239--266

\bibitem{DBLP:conf/aaai/LandwehrPRF06}
Landwehr, N., Passerini, A., De~Raedt, L., Frasconi, P.:
\newblock k{FOIL}: Learning simple relational kernels.
\newblock In: Proceedings of AAAI06, AAAI Press (2006)

\bibitem{Richardson06}
Richardson, M., Domingos, P.:
\newblock Markov logic networks.
\newblock Machine Learning \textbf{62} (2006)  107--136

\bibitem{Kramer:2001:PAR:567222.567236}
Kramer, S., Lavra\v{c}, N., Flach, P.:
\newblock Relational data mining.
\newblock Springer (2000)  262--286

\bibitem{ndm11ismis}
Di~Mauro, N., Basile, T.M., Ferilli, S., Esposito, F.:
\newblock Optimizing probabilistic models for relational sequence learning.
\newblock In: 19th International Symposium on Methodologies for Intelligent
  Systems. Volume 6804 of LNAI., Springer (2011)  240--249

\bibitem{Ho:1998:RSM:284980.284986}
Ho, T.K.:
\newblock The random subspace method for constructing decision forests.
\newblock IEEE Trans. Pattern Anal. Mach. Intell. \textbf{20} (1998)  832--844

\bibitem{DBLP:conf/aaai/LandwehrKR05}
Landwehr, N., Kersting, K., De~Raedti, L.:
\newblock n{FOIL}: Integrating na\"{\i}ve bayes and {FOIL}.
\newblock In: Proceedings of AAAI05, AAAI Press / The MIT Press (2005)
  795--800

\bibitem{DBLP:conf/uai/LangleyS94}
Langley, P., Sage, S.:
\newblock Induction of selective bayesian classifiers.
\newblock In de~M{\'a}ntaras, R.L., Poole, D., eds.: Proceedings of the Tenth
  Annual Conference on Uncertainty in Artificial Intelligence. (1994)  399--406

\bibitem{DBLP:conf/nips/WestonMCPPV00}
Weston, J., Mukherjee, S., Chapelle, O., Pontil, M., Poggio, T., Vapnik, V.:
\newblock Feature selection for svms.
\newblock In Leen, T.K., Dietterich, T.G., Tresp, V., eds.: Advances in Neural
  Information Processing Systems 13. (2000)  668--674

\bibitem{sls04}
Hoos, H., St{\"u}tzle, T.:
\newblock Stochastic Local Search: Foundations \& Applications.
\newblock Morgan Kaufmann Publishers Inc., San Francisco, CA, USA (2004)

\bibitem{feo95}
Feo, T., Resende, M.:
\newblock Greedy randomized adaptive search procedures.
\newblock Journal of Global Optimization \textbf{6} (1995)  109--133

\bibitem{Kramer01}
Kramer, S., Raedt, L.D.:
\newblock Feature construction with version spaces for biochemical
  applications.
\newblock In: Proceedings of the 18th International Conference on Machine
  Learning.
\newblock Morgan Kaufmann Publishers Inc. (2001)  258--265

\bibitem{ferilli03ismis}
Ferilli, S., {Di~Mauro}, N., Basile, T., Esposito, F.:
\newblock $\theta$-subsumption and resolution: A new algorithm.
\newblock In Zhong, N., Ra{\'s}, Z.W., Tsumoto, S., Suzuki, E., eds.:
  Foundations of Intelligent Systems. Volume 2871 of LNCS., Springer Verlag
  (2003)  384--391

\bibitem{Rokach:2010:EC:1713727.1713730}
Rokach, L.:
\newblock Ensemble-based classifiers.
\newblock Artif. Intell. Rev. \textbf{33} (2010)  1--39

\bibitem{Singla:2005:DTM:1619410.1619472}
Singla, P., Domingos, P.:
\newblock Discriminative training of markov logic networks.
\newblock In: Proceedings of AAAI05, AAAI Press (2005)  868--873

\bibitem{Srinivasan:1996:TMS:241085.241105}
Srinivasan, A., Muggleton, S.H., Sternberg, M.J.E., King, R.D.:
\newblock Theories for mutagenicity: a study in first-order and feature-based
  induction.
\newblock Artif. Intell. \textbf{85} (1996)  277--299

\bibitem{king95}
King, R.D., Sternberg, M.J., Srinivasan, A.:
\newblock Relating chemical activity to structure: An examination of ilp
  successes.
\newblock New Generation Computing \textbf{13} (1995)  411--433

\bibitem{DBLP:conf/icml/KokD10}
Kok, S., Domingos, P.:
\newblock Learning markov logic networks using structural motifs.
\newblock In F{\"u}rnkranz, J., Joachims, T., eds.: Proceedings of the 27th
  International Conference on Machine Learning. (2010)  551--558

\bibitem{Kok:2009:LML:1553374.1553440}
Kok, S., Domingos, P.:
\newblock Learning markov logic network structure via hypergraph lifting.
\newblock In: Proceedings of the 26th Annual International Conference on
  Machine Learning, ACM (2009)  505--512

\bibitem{DBLP:journals/ml/LandwehrPRF10}
Landwehr, N., Passerini, A., De~Raedt, L., Frasconi, P.:
\newblock Fast learning of relational kernels.
\newblock Machine Learning \textbf{78} (2010)  305--342

\end{thebibliography}

\end{document}